\documentclass[letterpaper]{article} 
\usepackage[preprint]{aaai2027}  
\usepackage[hyphens]{url}  
\usepackage{graphicx} 
\usepackage{natbib}  
\usepackage{caption} 
\usepackage{algorithm}
\usepackage{algorithmic}
\usepackage{amsmath}
\usepackage{amssymb}
\usepackage{tikz}
\usetikzlibrary{arrows.meta,positioning,shapes.geometric}

\usepackage{newfloat}
\usepackage{listings}
\DeclareCaptionStyle{ruled}{labelfont=normalfont,labelsep=colon,strut=off} 
\floatstyle{ruled}
\newfloat{listing}{tb}{lst}{}
\floatname{listing}{Listing}

\usepackage{booktabs}

\usepackage{multirow}
\usepackage[table]{xcolor}
\title{Self-Supervised Skill Optimization}

\author {
\normalsize
    Siran Peng\textsuperscript{\rm 1,2},\;
    Cuiyu Yang\textsuperscript{\rm 3,4},\;
    Tianyu Fu\textsuperscript{\rm 3},\;
    Tianshuo Zhang\textsuperscript{\rm 2,1},\;
    Haoyuan Zhang\textsuperscript{\rm 2,1},\;
    Weisong Zhao\textsuperscript{\rm 5},\\
    Anyang Su\textsuperscript{\rm 3},\;
    Minghui Wu\textsuperscript{\rm 3},\;
    Huiying Li\textsuperscript{\rm 4},\;
    Xiangyu Zhu\textsuperscript{\rm 1,2},\;
    Chenxu Zhao\textsuperscript{\rm 3*},\;
    Zhen Lei\textsuperscript{\rm 1,2,6*}
}
\affiliations{
    \textsuperscript{\rm 1}MAIS, CASIA,\;
    \textsuperscript{\rm 2}SAI, UCAS,\;
    \textsuperscript{\rm 3}Mininglamp Technology,\;
    \textsuperscript{\rm 4}JLU,\;
    \textsuperscript{\rm 5}IIE, CAS,\;
    \textsuperscript{\rm 6}SCSE, FIE, M.U.S.T \\
    \{pengsiran2023, zhen.lei\}@ia.ac.cn,\; zhaochenxu@mininglamp.com \\
    \textsuperscript{*}Corresponding author
}

\begin{document}
\definecolor{ssoblue}{RGB}{220,232,255}
\definecolor{rowgray}{RGB}{238,238,238}
\newcommand{\ssohighlight}[1]{\cellcolor{ssoblue}\strut #1}

\maketitle

\begin{abstract}
Agent skills provide frozen large language model (LLM) agents with reusable procedural guidance, and recent work shows that such skills can be optimized with ground-truth (GT) feedback. Many applications, however, lack GT labels, task scores, rewards, or reliable task-specific evaluators. We therefore introduce Self-Supervised Skill Optimization (SSO), a comparative framework that learns a reusable skill from unlabeled task instances alone. At each step, SSO runs the current skill on an unlabeled batch, uses a subset of the resulting executions to generate complete skill probes, and runs the probes on the same batch. An LLM judge compares the resulting answers, trajectories, artifacts, or terminal states. A separate behavior extractor identifies behavioral differences without seeing the judge's decisions. SSO uses these decisions to aggregate evidence for and against the observed behaviors across instances. It then ranks the behaviors by the resulting evidence and renders a new complete skill from the highest-ranked behaviors. The update is accepted only if the new skill outperforms the current one on an unlabeled validation set. SSO outperforms existing GT-free prompt optimizers on both closed-ended and open-ended tasks. On closed-ended benchmarks, it approaches and sometimes exceeds the strongest GT-based skill optimizer without using any GT feedback.
\end{abstract}

\section{Introduction}
Large language model (LLM) agents increasingly tackle tasks through multistep reasoning, tool use, and environment interaction~\cite{yao2023react,schick2023toolformer,wang2024voyager}. Such tasks require agents to coordinate decisions over extended executions, increasing the value of reusable procedural guidance. Agent skills provide such guidance to frozen agents by encoding procedures, policies for tool use, domain conventions, and output constraints~\cite{li2026skillsbench,jiang2026agenticskills}. In this work, a skill takes the form of a persistent document written in natural language. It can be inspected, edited, and reused across unseen task instances without changing model weights~\cite{ni2026trace2skill}. These properties make the skill document a natural target for optimization.
  
\begin{figure}[t]
	\begin{center}
		\begin{minipage}{1\linewidth}
			{\includegraphics[width=0.91\linewidth]{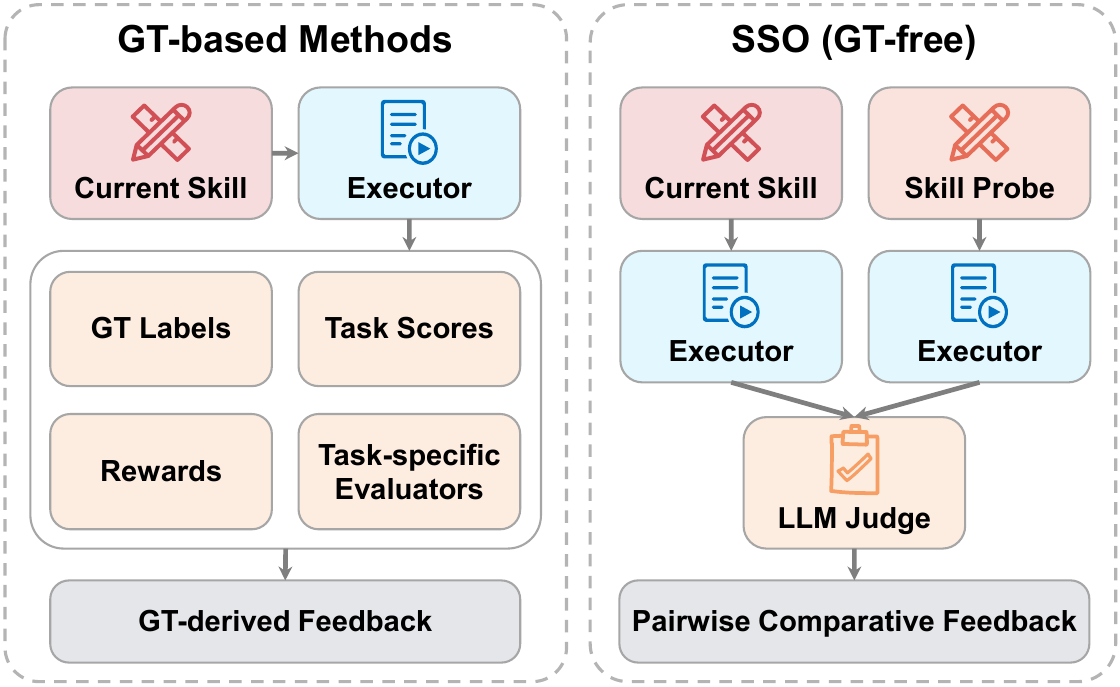}}
			\centering
		\end{minipage}
	\end{center}
	\caption{\textbf{GT-based skill optimization and SSO.} Existing methods use GT feedback to guide skill updates and selection. SSO instead derives optimization feedback from pairwise comparisons of observable executions on unlabeled task instances and uses no GT feedback during optimization. \label{fig:overview}}
\end{figure}
  
Recent work constructs and refines skills from execution trajectories, failures, and domain resources~\cite{ni2026trace2skill,alzubi2026evoskill,liu2026skillforge,shen2026skillfoundry}. SkillOpt~\cite{yang2026skillopt} directly treats a single skill document as the trainable state of a frozen agent. It converts scored rollouts into bounded edits and accepts an update only when it improves performance on a separate validation set. Its results show that reusable skills can be optimized across models, benchmarks, and execution harnesses. However, both its update directions and acceptance decisions rely on ground-truth (GT) feedback from task scores or task-specific evaluators.

Such feedback is often unavailable in practice. Open-ended tasks may have no reference answer, enterprise tasks may expose only source documents and execution traces, and interactive or artifact-producing tasks may lack a reliable task-specific evaluator. Collecting labels or building such evaluators can also diminish the practical benefit of adapting a frozen agent through a skill document. We therefore ask whether a reusable agent skill can be optimized from unlabeled task instances alone. We call this problem GT-free skill optimization and introduce Self-Supervised Skill Optimization (SSO), a comparative framework for addressing it. SSO does not use GT labels, task scores, rewards, or task-specific evaluators to update or select the skill. As shown in Figure~\ref{fig:overview}, it instead derives optimization feedback from pairwise comparisons of the agent's observable executions.

GT-free prompt optimization provides an important precedent, showing that pairwise comparisons can support prompt refinement without GT feedback~\cite{xiang2025spo,nair2025tournament,wu2025pdo,peng2026upa}. These methods use relative output quality to rank or select complete prompt candidates. This candidate-level treatment is natural when optimization is driven by response-level preferences. Its effectiveness, however, remains unclear in agentic settings that require extended executions, tool use, environment interaction, or artifact manipulation. Complex skill executions expose behavioral differences that can be compared across candidates and tasks. This motivates us to move beyond candidate-level selection and instead aggregate behavior-level evidence across comparisons to render a new skill.
  
SSO realizes this behavior-level approach through a comparative optimization loop over a single skill. At each step, SSO runs the current skill on an unlabeled batch, uses a subset of the resulting executions to generate complete skill probes, and runs the probes on the same batch. An LLM judge compares each probe execution with the corresponding execution under the current skill. A separate behavior extractor identifies behavioral differences without seeing the judge's decisions. SSO clusters semantically equivalent behaviors and uses the pairwise decisions to aggregate evidence for and against the observed behaviors across instances. When a decisive comparison can be linked to one or more observed behaviors, SSO distributes one unit of evidence equally among them. This normalization prevents comparisons involving more behaviors from exerting greater overall influence. SSO ranks the behaviors by the strength of the aggregated evidence and uses the highest-ranked behaviors to render a new complete skill. It accepts the new skill only if it outperforms the current skill on an unlabeled validation set. The probes serve only to expose alternative behaviors and are never accepted directly. The contributions of this paper are as follows. 
  
\begin{itemize}
\item We formalize GT-free skill optimization as the problem of optimizing a reusable skill for a frozen LLM agent from unlabeled task instances alone, without using GT feedback to generate, rank, or accept updates.

\item We introduce SSO, a comparative framework for improving reusable skills from agent executions on unlabeled task instances. It translates execution-level preferences into evidence for and against observed behaviors. It then ranks the behaviors by the resulting evidence and uses the highest-ranked behaviors to render a new complete skill.
  
\item We systematically evaluate SSO on closed-ended and open-ended tasks across multiple LLMs. SSO outperforms existing GT-free prompt optimizers in both settings. On closed-ended benchmarks, it approaches and sometimes exceeds the strongest GT-based skill optimizer.
\end{itemize}

\section{Related Work}


\subsection{Automatic Prompt Optimization}
Prompting techniques such as chain-of-thought (CoT) show that carefully designed prompts can improve reasoning performance~\cite{wei2022cot,NEURIPS2022_8bb0d291}. Automatic prompt optimization formulates prompt refinement as a search problem over textual instructions. APE~\cite{zhou2023ape} uses LLMs to generate and select candidate instructions, while ProTeGi~\cite{pryzant2023protegi} uses textual error feedback to propose prompt revisions. Subsequent methods guide this process with score histories, structured planning, or population-based evolution~\cite{yang2024opro,wang2024promptagent,fernando2024promptbreeder}. More recently, TextGrad~\cite{yuksekgonul2025textgrad} propagates textual feedback through compound AI systems, while GEPA~\cite{agrawal2025gepa} uses execution trajectories and natural-language reflection to optimize prompts and language programs.

Most prompt optimizers assume access to labeled examples, reference-based metrics, or task scores. Recent GT-free prompt optimizers instead compare outputs produced by competing prompts and use the resulting preferences to guide prompt refinement. SPO~\cite{xiang2025spo} greedily retains prompt revisions preferred by an LLM judge. PDO~\cite{wu2025pdo} casts label-free prompt selection as a dueling-bandit problem. UPA~\cite{peng2026upa} combines pairwise-guided tree search with uncertainty-aware filtering and global tournament selection. Beyond prompt optimization, pairwise LLM evaluation is widely used to compare open-ended responses, although it can be sensitive to factors such as response order~\cite{zheng2023mtbench}. Existing GT-free prompt optimizers have been evaluated primarily on response-level benchmarks. Their effectiveness in agentic settings that require extended executions, tool use, environment interaction, or artifact manipulation therefore remains unclear. Because they operate on response-level preferences, these methods treat complete prompt candidates as the unit of optimization. Agentic skill executions expose additional behavioral structure. Current methods do not study how evidence from this structure can be combined across probes and tasks to construct a skill beyond the compared candidates.

\begin{figure*}[t]
	\begin{center}
		\begin{minipage}{1\linewidth}
			{\includegraphics[width=0.89\linewidth]{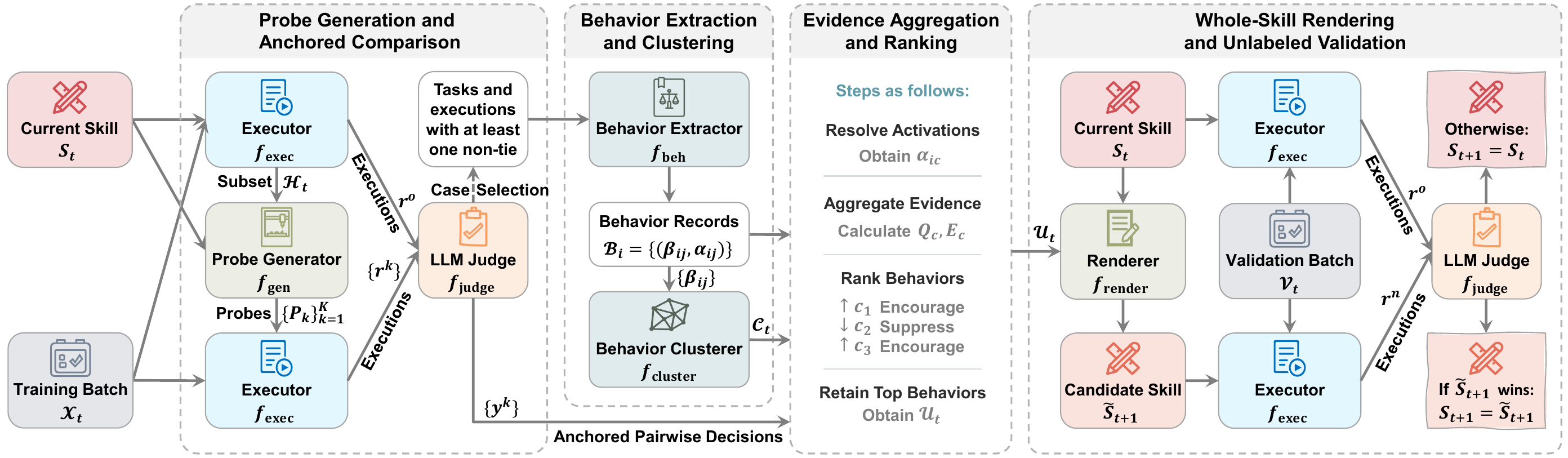}}
			\centering
		\end{minipage}
	\end{center}
	\caption{\textbf{The SSO optimization loop.} SSO executes the current skill and $K$ complete probes on the same unlabeled training batch. An LLM judge then produces anchored pairwise decisions. A behavior extractor identifies observable differences without seeing the decisions, and a separate clusterer groups equivalent descriptions. SSO combines the decisions with behavior activations to rank behaviors and render a new skill, accepting it only if it wins against the current skill on a validation batch. \label{fig:method}}
\end{figure*}
  
\subsection{Agent Skills and Skill Optimization}
Recent work has established agent skills as persistent, reusable artifacts that extend beyond atomic tool calls. SkillsBench~\cite{li2026skillsbench} studies how skills affect agent performance, while the SoK on agent skills~\cite{jiang2026agenticskills} systematizes their representations, design patterns, and lifecycle. Building on this abstraction, one line of work focuses on acquiring and organizing reusable skills. Voyager~\cite{wang2024voyager}, Trace2Skill~\cite{ni2026trace2skill}, AutoSkill~\cite{liu2026autoskill}, AutoRefine~\cite{shen2026autorefine}, and SkillFoundry~\cite{shen2026skillfoundry} construct skills from embodied experience, execution trajectories, user interactions, or heterogeneous external resources. These systems primarily study skill acquisition, distillation, and library maintenance rather than iterative optimization of a single skill.

A second line of work iteratively refines or optimizes skill artifacts using task outcomes. EvoSkill~\cite{alzubi2026evoskill} combines failure analysis with validation scores, SkillForge~\cite{liu2026skillforge} uses execution failures to refine skills for cloud-support tasks, and CoEvoSkills~\cite{chen2026coevoskills} co-evolves skills with a surrogate verifier while still consulting a GT oracle. SkillGen~\cite{chen2026skillgen} induces a single auditable skill from successful and failed trajectories and evaluates candidate skills through verified repairs and regressions, while HDSO~\cite{li2026hdso} uses observed success or failure to validate skill hypotheses through paired control and treatment executions. Among these methods, SkillOpt~\cite{yang2026skillopt} presents a general framework for GT-based optimization of a single reusable skill. It treats the skill document as the trainable state of a frozen agent, converts scored rollouts into bounded edits, and accepts an update only when it improves performance on a validation set. Together, these methods show that reusable skills can be optimized effectively, but GT feedback still informs their update directions, acceptance decisions, or both.

Recent studies have explored adjacent forms of skill learning and evaluation without GT feedback~\cite{sun2026ctx2skill,huang2026datacope,zhao2026skillaudit,wu2026skillaxe,li2026skillscoach,li2026openskill}. These works focus on context-specific skills, specialized domains, task-local analysis, or skill libraries. None provides a systematic cross-domain study of GT-free optimization for a single reusable skill over an unlabeled task distribution. We address this gap with SSO, a general comparative framework evaluated across closed-ended and open-ended tasks.  

\section{Methodology}
  
\subsection{Problem Setup and Overview}
Let $\mathcal{D}_{\mathrm{tr}}=\{x_i\}$ be an unlabeled training set of task instances, and let $\mathcal{D}_{\mathrm{val}}$ be an unlabeled validation set disjoint from $\mathcal{D}_{\mathrm{tr}}$. Let $f_{\mathrm{exec}}$ denote the fixed execution function induced by a frozen LLM, and let $S$ be a complete reusable skill document. Executing $S$ on task $x$ produces an observable execution $r$:
\begin{equation}
r=f_{\mathrm{exec}}(S,x).
\end{equation}
Here, $r$ may include an answer, a trajectory, an artifact, or a terminal state, depending on the task. A trajectory may contain tool calls, observations, or environment actions.
GT-free skill optimization assumes that no GT labels, task scores, rewards, or task-specific evaluators are available during optimization. Given $\mathcal{D}_{\mathrm{tr}}$, $\mathcal{D}_{\mathrm{val}}$, $f_{\mathrm{exec}}$, and an initial skill $S_0$, the goal is to learn a reusable skill $S^*$ that improves expected performance relative to $S_0$ on unseen tasks from the same domain, without using GT feedback to generate or select updates. SSO is our solution for this problem.

Figure~\ref{fig:method} illustrates the SSO optimization loop. At each step, SSO runs the current skill on an unlabeled batch and uses a subset of the resulting executions to generate $K$ complete skill probes. It then runs every probe on the same batch, producing alternative executions on matched task instances. A fixed LLM judge compares each probe execution with the corresponding execution under the current skill. Rather than reducing these comparisons to a ranking of probes, SSO uses a separate behavior extractor to identify behavioral differences without seeing the judge's decisions. A separate clusterer groups semantically equivalent behaviors. SSO then aggregates evidence for and against them across probes and tasks and ranks them by the resulting scores. Finally, SSO renders a new complete skill from the highest-ranked behaviors and accepts it only if it outperforms the current skill on an unlabeled validation set. Algorithm~\ref{alg:sso} summarizes the full procedure of SSO.

\begin{algorithm}[!t]
\caption{Self-Supervised Skill Optimization (SSO)}
\label{alg:sso}
\begin{algorithmic}[1]
\REQUIRE $S_0$, $\mathcal{D}_{\mathrm{tr}}$, $\mathcal{D}_{\mathrm{val}}$, $K$, $U$, and optimization steps $T$
\ENSURE Optimized skill $S^*$
\FOR{$t=0,\ldots,T-1$}
    \STATE \textcolor{blue}{\textsc{// Probe Generation and Anchored Comparison}}
    \STATE Select $\mathcal{X}_t\subset\mathcal{D}_{\mathrm{tr}}$
    \STATE $r_i^o\leftarrow f_{\mathrm{exec}}(S_t,x_i)$
    for each $x_i\in\mathcal{X}_t$
    \STATE Select history subset $\mathcal{H}_t$
    \STATE $\{P_k\}_{k=1}^{K}\leftarrow
    f_{\mathrm{gen}}(S_t,\mathcal{H}_t)$

    \FOR{$k=1,\ldots,K$}
        \STATE $r_i^k\leftarrow f_{\mathrm{exec}}(P_k,x_i)$
        for each $x_i\in\mathcal{X}_t$
        \STATE $y_i^k\leftarrow
        f_{\mathrm{judge}}(x_i,r_i^k,r_i^o)$
        for each $x_i\in\mathcal{X}_t$
    \ENDFOR

    \STATE \textcolor{blue}{\textsc{// Behavior Extraction and Clustering}}
    \FOR{each $x_i\in\mathcal{X}_t$ with $\exists k,\ y_i^k\neq0$}
        \STATE $\{(\beta_{ij},\alpha_{ij})\}_{j=1}^{J_i}\leftarrow
        f_{\mathrm{beh}}(x_i,r_i^o,\{r_i^k\}_{k=1}^{K})$
    \ENDFOR
    \STATE $\mathcal{C}_t\leftarrow
    f_{\mathrm{cluster}}(\{\beta_{ij}\}_{i,j})$

    \STATE \textcolor{blue}{\textsc{// Evidence Aggregation and Ranking}}
    \STATE Resolve $\alpha_{ic}(r)$ from the clustered records
    \STATE $Q_c,E_c\leftarrow0$ for each $c\in\mathcal{C}_t$

    \FOR{each $(i,k)$ with $y_i^k\neq0$}
        \STATE Form the changed-cluster set $\mathcal{A}_i^k$
        \FOR{each $c\in\mathcal{A}_i^k$}
            \STATE $d_{ic}^k\leftarrow
            \alpha_{ic}(r_i^k)-\alpha_{ic}(r_i^o)$
            \STATE $e_{ic}^k\leftarrow
            y_i^k d_{ic}^k/|\mathcal{A}_i^k|$
            \STATE $Q_c\leftarrow Q_c+e_{ic}^k$
            \STATE $E_c\leftarrow E_c+|e_{ic}^k|$
        \ENDFOR
    \ENDFOR
    \STATE Set intents from $\operatorname{sign}(Q_c)$ and discard $Q_c=0$
    \STATE Rank the rest by $|Q_c|$, $E_c$, and contributing-task count
    \STATE Retain up to $U$ highest-ranked clusters as $\mathcal{U}_t$

    \STATE \textcolor{blue}{\textsc{// Whole-skill Rendering and Validation}}
    \IF{$\mathcal{U}_t=\varnothing$}
        \STATE $S_{t+1}\leftarrow S_t$ and \textbf{continue}
    \ENDIF

    \STATE $\widetilde{S}_{t+1}\leftarrow
    f_{\mathrm{render}}(S_t,\mathcal{U}_t)$
    \STATE Sample $\mathcal{V}_t\subseteq\mathcal{D}_{\mathrm{val}}$

    \FOR{each $x_m\in\mathcal{V}_t$}
        \STATE Obtain $r_m^o$ and $r_m^n$ using $f_{\mathrm{exec}}$
        \STATE $v_m\leftarrow
        f_{\mathrm{judge}}(x_m,r_m^n,r_m^o)$
    \ENDFOR
    \STATE Count candidate wins $W_t$ and losses $L_t$
    \STATE $S_{t+1}\leftarrow\widetilde{S}_{t+1}$
    if $W_t>L_t$; otherwise $S_t$
\ENDFOR
\STATE $S^*\leftarrow S_T$
\RETURN $S^*$
\end{algorithmic}
\end{algorithm}

\subsection{Probe Generation and Anchored Comparison}
At step $t$, SSO begins by selecting an unlabeled training batch $\mathcal{X}_t\subset\mathcal{D}_{\mathrm{tr}}$. Executing the current skill $S_t$ on each task $x_i\in\mathcal{X}_t$ yields an observable execution $r_i^{o}$:
\begin{equation}
r_i^{o}=f_{\mathrm{exec}}(S_t,x_i),\quad x_i\in\mathcal{X}_t,
\end{equation}
which serves as the anchor for subsequent comparisons. To produce skill probes for exposing alternative behaviors, SSO employs an LLM-based probe generator $f_{\mathrm{gen}}$. Conditioning on the current skill $S_t$ and a subset $\mathcal{H}_t$ of its task-execution history, $f_{\mathrm{gen}}$ jointly generates $K$ skill probes:
\begin{equation}
\{P_k\}_{k=1}^{K}=f_{\mathrm{gen}}(S_t,\mathcal{H}_t),\quad
\mathcal{H}_t\subseteq\left\{(x_i,r_i^{o})\mid x_i\in\mathcal{X}_t\right\}.
\end{equation}
Each $P_k$ is a complete skill document rather than an isolated edit. A probe may revise multiple parts of $S_t$, and is constrained to be distinct from both $S_t$ and its peer probes.
  
Each probe is then executed on the same training batch:
\begin{equation}
r_i^{k}=f_{\mathrm{exec}}(P_k,x_i),\quad x_i\in\mathcal{X}_t,\quad k=1,\ldots,K.
\end{equation}
Here, $r_i^{k}$ denotes the execution of the $k$-th probe on task $x_i$.
SSO uses a fixed comparative judge $f_{\mathrm{judge}}$, implemented by a frozen LLM, to compare each probe execution with its corresponding anchor:
\begin{equation}
y_{i}^{k}=f_{\mathrm{judge}}\!\left(x_i,r_i^{k},r_i^{o}\right)\in\{-1,0,+1\},
\end{equation}
where $y_{i}^{k}=+1$ indicates that the probe execution is preferred, $y_{i}^{k}=-1$ indicates that the anchor execution is preferred, and $y_{i}^{k}=0$ denotes a tie. For each task, all probe executions are compared against the same current-skill execution, so no comparisons among probes are required. To mitigate position bias, we randomize the presentation order of the executions and map the resulting decision back to their probe and anchor identities before assigning $y_{i}^{k}$. Rather than using these comparisons only to select one of the generated probes, SSO retains their task-level evidence to construct a new skill that need not match any individual probe.

\subsection{Behavior Extraction and Clustering}
SSO next represents observable execution differences as behaviors, providing a common unit for aggregating evidence across probes and tasks. Semantically equivalent behaviors can then accumulate evidence for and against them across comparisons. To prevent the pairwise outcomes from biasing which differences are reported or how they are described, the behavior extractor does not see the judge decisions.
For each task with at least one decisive comparison (i.e., at least one non-tie), an LLM-based behavior extractor $f_{\mathrm{beh}}$ receives the task, the current execution, and all probe executions:
\begin{equation}
\mathcal{B}_i=\{(\beta_{ij},\alpha_{ij})\}_{j=1}^{J_i}=f_{\mathrm{beh}}\!\left(x_i,r_i^{o},\{r_i^{k}\}_{k=1}^{K}\right).
\end{equation}
Here, $\mathcal{B}_i$ is the set of behavior records extracted for task $x_i$, $J_i=|\mathcal{B}_i|$, and $j$ indexes these records. Each record pairs a behavior description $\beta_{ij}$ with a mapping $\alpha_{ij}$ that assigns an activation state to each input execution:
\begin{equation}
\alpha_{ij}(r)\in\{1,0,\bot\}.
\end{equation}
The values $1$, $0$, and $\bot$ indicate that the behavior is present, absent, or unclear, respectively. The extractor may return multiple behaviors when the input executions contain multiple observable differences.

The behavior descriptions extracted from these task instances are then grouped by an LLM-based clusterer $f_{\mathrm{cluster}}$:
\begin{equation}
\mathcal{C}_t=f_{\mathrm{cluster}}\!\left(\{\beta_{ij}\}_{i,j}\right).
\end{equation}
The clusterer receives only the behavior descriptions, without activation states or pairwise decisions, and partitions semantically equivalent descriptions into clusters. It groups descriptions only when they express the same observable behavior; related but distinct behaviors remain separate. Each cluster is represented by one of its original descriptions rather than a newly generated summary, ensuring that clustering does not alter the meaning of the behavior.
  
\subsection{Evidence Aggregation and Ranking}
After clustering, SSO combines the pairwise decisions with behavior activations. For cluster $c\in\mathcal{C}_t$, let $\alpha_{ic}(r)\in\{1,0,\bot\}$ denote the activation of its behavior in execution $r$ for task $x_i$. For each $r$, $\alpha_{ic}(r)$ is the common clear activation assigned by the task-local behavior records in cluster $c$. For a given execution $r$, $\alpha_{ic}(r)=\bot$ if no such record exists, the activation is unclear, or the records disagree. For each comparison, SSO first identifies the clusters whose behaviors clearly differ between the probe and anchor executions:
\begin{equation}
  \mathcal{A}_i^k=
  \left\{c\in\mathcal{C}_t\,\middle|\,
  \substack{
  \alpha_{ic}(r_i^k),\alpha_{ic}(r_i^o)\in\{0,1\},\\
  \alpha_{ic}(r_i^k)\neq\alpha_{ic}(r_i^o)
  }
  \right\}.
\end{equation}
For each $c\in\mathcal{A}_i^k$, the activation difference is:
\begin{equation}
d_{ic}^k=\alpha_{ic}(r_i^k)-\alpha_{ic}(r_i^o)\in\{-1,+1\}.
\end{equation}
Here, $d_{ic}^k=+1$ means that the behavior appears in the probe execution, while $d_{ic}^k=-1$ means that it disappears. For a decisive comparison with $\mathcal{A}_i^k\neq\varnothing$, SSO assigns evidence to each changed behavior cluster:
\begin{equation}
e_{ic}^k=\frac{y_i^k d_{ic}^k}{|\mathcal{A}_i^k|},
\quad c\in\mathcal{A}_i^k.
\end{equation}
All other cluster-comparison pairs receive zero evidence. Because $y_i^k$ identifies the preferred execution and $d_{ic}^k$ identifies where the behavior appears, a behavior present only in the preferred execution receives positive evidence, whereas one present only in the non-preferred execution receives negative evidence. For example, if a preferred probe introduces two changed behaviors, each receives $+1/2$ evidence. If the anchor is preferred instead, each receives $-1/2$.
Each decisive comparison with at least one clear behavioral difference contributes one normalized unit of absolute evidence:
\begin{equation}
\sum_{c\in\mathcal{A}_i^k}|e_{ic}^k|=1.
\end{equation}
This normalization gives every such comparison the same total influence regardless of how many behaviors differ between its executions. A decisive comparison with no clear behavioral difference is left unattributed and does not affect any cluster score. The signed score and total evidence mass for cluster $c$ are:
\begin{equation}
Q_c=\sum_{i,k}e_{ic}^k,\quad
E_c=\sum_{i,k}|e_{ic}^k|.
\end{equation}
For example, evidence values $+1$, $+1$, and $-1$ give $Q_c=1$ and $E_c=3$: the net direction is positive, while the total evidence mass is three.

The sign of $Q_c$ determines the update direction:
\begin{equation}
\operatorname{intent}(c)=
\begin{cases}
\mathrm{encourage}, & Q_c>0,\\
\mathrm{suppress}, & Q_c<0,\\
\mathrm{neutral}, & Q_c=0.
\end{cases}
\end{equation}
Clusters with $Q_c=0$ are discarded. The remaining clusters are ranked by descending $|Q_c|$, with larger $E_c$ and evidence from more distinct tasks used as successive tie-breakers. Any remaining ties are resolved deterministically. Given an update budget $U$, SSO retains up to $U$ highest-ranked clusters as the behavior units $\mathcal{U}_t$. Each unit contains the representative behavior description, its update direction, its signed score $Q_c$, and the evidence for and against the behavior.

\begin{table*}[!t]
  \centering
  \caption{
  Closed-ended evaluation results (\%) across six benchmarks and three target models, with an additional GPT-5.5 evaluation under the Codex harness. Entries are grouped by whether GT is used during skill construction, optimization, or selection. Results for SPO and PDO are reproduced, whereas results for other baselines are reported from \cite{yang2026skillopt}. The Codex average is computed over five benchmarks because ALFWorld requires a separate persistent embodied environment.
  }
  \label{tab:main}
  \setlength{\tabcolsep}{5.5pt}
  \renewcommand{\arraystretch}{1.05}

  \begin{tabular}{ll ccccccc}
  \toprule
  \textbf{Supervision} &
  \textbf{Method} &
  \textbf{SearchQA} &
  \textbf{Spreadsheet} &
  \textbf{OfficeQA} &
  \textbf{DocVQA} &
  \textbf{LiveMath} &
  \textbf{ALFWorld} &
  \textbf{Average} \\
  \midrule

  \rowcolor{rowgray}
  \multicolumn{9}{c}{\textbf{Target model: GPT-5.5}} \\
  \multirow{5}{*}{GT-based}
   & Human skill  & 81.8 & 72.9 & 66.9 & 90.1 & 38.4 & 91.8 & 73.7 \\
   & Trace2Skill  & 82.4 & 49.6 & 65.7 & 90.6 & 52.0 & 87.3 & 71.3 \\
   & TextGrad     & 81.4 & 41.1 & 42.0 & 87.2 & 49.2 & 82.8 & 64.0 \\
   & GEPA         & 84.8 & 73.6 & 63.9 & 89.1 & 43.2 & 85.8 & 73.4 \\
   & SkillOpt     & 87.3 & 80.7 & 72.1 & 91.2 & 66.9 & 95.5 & 82.3 \\
  \cmidrule(lr){1-9}
  \multirow{4}{*}{GT-free}
   & No skill & 77.7 & 41.8 & 33.1 & 78.8 & 37.6 & 83.6 & 58.8 \\
   & SPO      & 81.8 & 43.9 & 62.2 & 86.4 & 54.0 & 88.1 & 69.4 \\
   & PDO      & 80.1 & 47.9 & 64.0 & 85.8 & 47.6 & 86.6 & 68.7 \\
   & \ssohighlight{\textbf{SSO}} & \ssohighlight{\textbf{83.9}} & \ssohighlight{\textbf{78.9}} & \ssohighlight{\textbf{66.9}} & \ssohighlight{\textbf{91.7}} & \ssohighlight{\textbf{62.1}} & \ssohighlight{\textbf{91.8}} & \ssohighlight{\textbf{79.2}} \\
  \midrule

  \rowcolor{rowgray}
  \multicolumn{9}{c}{\textbf{Target model: GPT-5.4-mini}} \\
  \multirow{5}{*}{GT-based}
   & Human skill  & 77.2 & 42.9 & 45.9 & 85.0 & 28.8 & 56.7 & 56.1 \\
   & Trace2Skill  & 78.6 & 40.7 & 20.9 & 88.5 & 32.8 & 82.8 & 57.4 \\
   & TextGrad     & 77.5 & 38.2 & 30.0 & 84.0 & 27.2 & 70.9 & 54.6 \\
   & GEPA         & 79.4 & 42.5 & 45.3 & 83.7 & 27.2 & 81.3 & 59.9 \\
   & SkillOpt     & 80.2 & 47.5 & 48.8 & 90.9 & 32.8 & 85.8 & 64.3 \\
  \cmidrule(lr){1-9}
  \multirow{4}{*}{GT-free}
   & No skill & 75.9 & 36.1 & 22.1 & 71.4 & 14.7 & 73.1 & 48.9 \\
   & SPO      & 78.0 & 40.7 & 45.9 & 85.6 & 29.8 & 79.1 & 59.9 \\
   & PDO      & 76.3 & 39.6 & 48.3 & 83.7 & 34.7 & 80.6 & 60.5 \\
   & \ssohighlight{\textbf{SSO}} & \ssohighlight{\textbf{78.8}} & \ssohighlight{\textbf{55.0}} & \ssohighlight{\textbf{49.4}} & \ssohighlight{\textbf{89.8}} & \ssohighlight{\textbf{37.9}} & \ssohighlight{\textbf{82.1}} & \ssohighlight{\textbf{65.5}} \\
  \midrule

  \rowcolor{rowgray}
  \multicolumn{9}{c}{\textbf{Target model: Qwen3.5-4B}} \\
  \multirow{5}{*}{GT-based}
   & Human skill  & 66.3 & 16.4 & 22.7 & 87.8 & 18.4 & 28.4 & 40.0 \\
   & Trace2Skill  & 68.5 & 19.3 & 16.3 & 88.0 & 27.2 & 64.9 & 47.4 \\
   & TextGrad     & 60.7 & 13.9 & 20.9 & 85.6 & 10.6 & 53.7 & 40.9 \\
   & GEPA         & 68.6 & 16.9 & 22.7 & 85.1 & 28.8 & 60.4 & 47.1 \\
   & SkillOpt     & 71.2 & 23.9 & 29.7 & 89.0 & 52.0 & 81.3 & 57.9 \\
  \cmidrule(lr){1-9}
  \multirow{4}{*}{GT-free}
   & No skill & 68.1 &  9.3 & 14.5 & 86.9 & 22.4 & 30.6 & 38.6 \\
   & SPO      & 66.8 & 16.1 & 24.4 & 87.7 & 33.1 & 57.5 & 47.6 \\
   & PDO      & 62.8 & 17.9 & 22.1 & 88.0 & 28.2 & 56.7 & 46.0 \\
   & \ssohighlight{\textbf{SSO}} & \ssohighlight{\textbf{70.2}} & \ssohighlight{\textbf{21.8}} & \ssohighlight{\textbf{30.8}} & \ssohighlight{\textbf{89.6}} & \ssohighlight{\textbf{39.5}} & \ssohighlight{\textbf{64.9}} & \ssohighlight{\textbf{52.8}} \\
    \midrule
    \rowcolor{rowgray}
  \multicolumn{9}{c}{\textbf{Codex harness (target model: GPT-5.5)}} \\
  \multirow{3}{*}{GT-based}
   & Human skill & 84.1 & 50.7 & 40.0 & 88.8 & 48.8 & -- & 62.5 \\
   & EvoSkill    & 61.4 & 67.5 & 42.4 & 89.3 & 63.2 & -- & 64.8 \\
   & SkillOpt    & 87.3 & 85.0 & 51.1 & 92.2 & 78.4 & -- & 78.8 \\
  \cmidrule(lr){1-9}
  \multirow{2}{*}{GT-free}
   & No skill & 81.8 & 27.5 & 38.3 & 87.2 & 35.2 & -- & 54.0 \\
   & \ssohighlight{\textbf{SSO}} & \ssohighlight{\textbf{84.8}} & \ssohighlight{\textbf{72.1}} & \ssohighlight{\textbf{70.9}} & \ssohighlight{\textbf{91.4}} & \ssohighlight{\textbf{65.3}} & \ssohighlight{--} & \ssohighlight{\textbf{76.9}} \\
  \bottomrule
  \end{tabular}
\end{table*}
  
\subsection{Whole-Skill Rendering and Unlabeled Validation}
If $\mathcal{U}_t=\varnothing$, SSO does not generate a candidate and keeps the current skill unchanged. Otherwise, an LLM-based renderer $f_{\mathrm{render}}$ receives the current skill and the selected behaviors:
\begin{equation}
\widetilde{S}_{t+1}=f_{\mathrm{render}}(S_t,\mathcal{U}_t).
\end{equation}
The renderer outputs a new skill. It may add or strengthen an encouraged behavior and remove, weaken, or restrict a suppressed behavior, but it does not introduce an unsupported opposite. Because $\mathcal{U}_t$ aggregates evidence across probes and tasks, the candidate can combine supported behaviors from different probes and need not match any individual probe.

SSO then samples a validation batch $\mathcal{V}_t\subseteq\mathcal{D}_{\mathrm{val}}$ and executes the current and rendered skills on every task $x_m\in\mathcal{V}_t$:
\begin{equation}
r_m^{o}=f_{\mathrm{exec}}(S_t,x_m),\quad
r_m^{n}=f_{\mathrm{exec}}(\widetilde{S}_{t+1},x_m).
\end{equation}
Here, $r_m^{o}$ and $r_m^{n}$ denote the current-skill and candidate executions. The same comparative judge compares each candidate execution with its corresponding current-skill execution:
\begin{equation}
v_m=f_{\mathrm{judge}}\!\left(x_m,r_m^{n},r_m^{o}\right)\in\{-1,0,+1\}.
\end{equation}
Following the convention for $y_i^k$, $v_m=+1$, $-1$, or $0$ denotes a candidate win, candidate loss, or tie, respectively. As in the training comparisons, validation uses no GT feedback and applies the same presentation-order randomization and identity mapping. Let $W_t$ and $L_t$ denote the numbers of candidate wins and losses on $\mathcal{V}_t$. The update rule is:
\begin{equation}
S_{t+1}=
\begin{cases}
\widetilde{S}_{t+1}, & W_t>L_t,\\
S_t, & \mathrm{otherwise}.
\end{cases}
\end{equation}
Ties contribute to neither count, and the candidate is accepted only when it wins more comparisons than it loses.
  
\section{Experiments}
\subsection{Experimental Setup}
\paragraph{Benchmarks.}
We evaluate SSO on six closed-ended benchmarks and three open-ended multi-turn dialogue tasks. The closed-ended suite covers retrieval QA with SearchQA~\cite{dunn2017searchqa}, spreadsheet artifact manipulation with SpreadsheetBench~\cite{ma2024spreadsheetbench}, grounded enterprise QA with OfficeQA~\cite{opsahlOng2026officeqa}, multimodal document understanding with DocVQA~\cite{mathew2021docvqa}, advanced multiple-choice mathematics with LiveMathematicianBench~\cite{he2026livemathbench}, and embodied household decision making with ALFWorld~\cite{shridhar2021alfworld}. For open-ended evaluation, we use three tasks from MT-Bench-101~\cite{bai2024mt}: Context Memory (CM), Content Rephrasing (CR), and Proactive Interaction (PI). They respectively measure the model's perceptivity, adaptability, and interactivity. Additional dataset details are provided in the supplementary material.

\paragraph{Baselines.} 
For closed-ended tasks, we compare SSO with two categories of baselines. The first includes GT-based baselines: Human skill, Trace2Skill~\cite{ni2026trace2skill}, TextGrad~\cite{yuksekgonul2025textgrad}, GEPA~\cite{agrawal2025gepa}, EvoSkill~\cite{alzubi2026evoskill}, and SkillOpt~\cite{yang2026skillopt}. Human skill denotes the human-expert-written skill from \cite{yang2026skillopt}. The second includes No skill and two GT-free prompt optimizers, SPO~\cite{xiang2025spo} and PDO~\cite{wu2025pdo}. We adapt both prompt optimizers to optimize complete skills. For open-ended tasks, we compare SSO with No skill and the adapted SPO and PDO baselines.  

\paragraph{Metrics.} 
We report hard accuracy for SearchQA, OfficeQA, DocVQA, and LiveMathematicianBench, official task accuracy for SpreadsheetBench, and task success rate for ALFWorld. All metrics are reported as percentages, with unweighted averages across benchmarks. The Codex~\cite{openai2025codex} average is computed over the five applicable benchmarks because ALFWorld requires a separate persistent embodied environment. For MT-Bench-101, an LLM judge compares SSO with each baseline. We report the percentage of comparisons in which SSO is preferred as the pairwise win rate; values above 50 indicate that the judge favors SSO over the corresponding baseline overall. Details of the LLM judge are provided in the supplementary material.

\paragraph{Implementation Details.} 
We evaluate three frozen target models: GPT-5.5~\cite{openai2026gpt55}, GPT-5.4-mini~\cite{openai2026gpt54}, and Qwen3.5-4B~\cite{qwen2026qwen35}. The target model instantiates $f_{\mathrm{exec}}$ under the corresponding benchmark harness. All optimizer-side LLM modules ($f_{\mathrm{gen}}$, $f_{\mathrm{judge}}$, $f_{\mathrm{beh}}$, $f_{\mathrm{cluster}}$, and $f_{\mathrm{render}}$) use GPT-5.5. We initialize $S_0$ as the no-skill condition. SSO uses $K=3$ probes, $|\mathcal{H}_t|=8$ task-execution pairs for probe generation, and an update budget of $U=4$ behavior units. The optimization schedule varies by dataset: $T$ ranges from $8$ to $40$, $|\mathcal{X}_t|$ from $20$ to $40$, and $|\mathcal{V}_t|$ from $16$ to $40$. These schedules keep the optimization scale comparable to SkillOpt while accounting for dataset size. Complete schedules, prompts for all LLM modules, and other details are provided in the supplementary material.

\begin{table}[!t]
\footnotesize
\centering
\caption{
  Pairwise win rates (\%) of SSO against GT-free baselines on three tasks from MT-Bench-101 across three target models. CM, CR, and PI denote Context Memory, Content Rephrasing, and Proactive Interaction. Each entry reports the percentage of comparisons in which SSO is preferred; a value above 50 indicates that the judge favors SSO overall.
    }
  \label{tab:open}
  \setlength{\tabcolsep}{1.9pt}
  \renewcommand{\arraystretch}{1.05}
  \begin{tabular}{l ccc ccc ccc}
  \toprule
  \multirow{2}{*}{\textbf{Comparison}} &
  \multicolumn{3}{c}{\textbf{GPT-5.5}} &
  \multicolumn{3}{c}{\textbf{GPT-5.4-mini}} &
  \multicolumn{3}{c}{\textbf{Qwen3.5-4B}} \\
  \cmidrule(lr){2-4}
  \cmidrule(lr){5-7}
  \cmidrule(lr){8-10}
  & \textbf{CM} & \textbf{CR} & \textbf{PI}
  & \textbf{CM} & \textbf{CR} & \textbf{PI}
  & \textbf{CM} & \textbf{CR} & \textbf{PI} \\
  \midrule
  \textbf{SSO} vs.\ No skill
    & 58.3 & 75.0 & 54.2
    & 53.3 & 73.8 & 53.4
    & 55.0 & 82.5 & 56.2 \\

  \textbf{SSO} vs.\ SPO
    & 55.0 & 65.0 & 64.1
    & 50.8 & 71.3 & 84.0
    & 54.2 & 63.8 & 64.4 \\

  \textbf{SSO} vs.\ PDO
    & 52.5 & 66.3 & 58.8
    & 56.7 & 63.8 & 60.5
    & 58.3 & 75.0 & 70.3 \\
  \bottomrule
  \end{tabular}
\end{table}
  
\subsection{Performance}
\paragraph{Closed-ended Tasks.}
Table~\ref{tab:main} shows that SSO is the strongest GT-free method across all benchmark-model pairs. It improves the average over No skill by 20.4, 16.6, and 14.2 points with GPT-5.5, GPT-5.4-mini, and Qwen3.5-4B, respectively. With GPT-5.5, SSO outperforms SPO by 9.8 points and recovers 86.8\% of SkillOpt's gain without GT feedback. It also exceeds SkillOpt by 1.2 points with GPT-5.4-mini. Under the Codex harness, SSO improves over No skill by 22.9 points and trails SkillOpt by only 1.9 points. The optimized skills are provided in the supplementary material.

\paragraph{Open-ended Tasks.}
Table~\ref{tab:open} shows that SSO achieves win rates above 50 against every baseline across all tasks and target models. Averaged over tasks and models, its win rates are 62.4\% against No skill, 63.6\% against SPO, and 62.5\% against PDO. These results demonstrate the effectiveness of SSO in open-ended dialogue settings. GT-based methods are not included because these tasks lack GT labels or scores.
  
\subsection{Ablation Studies}
\begin{table}[t]
\centering
\footnotesize
\caption{Component ablations (\%) on three closed-ended benchmarks with GPT-5.4-mini as the target model.}
\label{tab:ablation}
\setlength{\tabcolsep}{1.8pt}
\renewcommand{\arraystretch}{1.05}
\begin{tabular}{lccc}
\hline
\textbf{Variant} & \textbf{Spreadsheet} & \textbf{DocVQA} & \textbf{LiveMath} \\
\hline
Direct Probe Selection    & 42.9 & 86.1 & 28.2 \\
w/o Behavior Clustering  & 51.1 & 88.0 & 33.1  \\
w/o Evidence Normalization  & 51.8 & 89.3 & 35.5  \\
Random Behavior Selection  & 46.1 & 85.3 & 33.1  \\
w/o Validation Gate      & 50.0 & 87.7 & 36.3 \\
\ssohighlight{\textbf{Full SSO}} & \ssohighlight{\textbf{55.0}} & \ssohighlight{\textbf{89.8}} & \ssohighlight{\textbf{37.9}} \\
\hline
\end{tabular}
\end{table}

\begin{figure}[t]
\begin{center}
		\begin{minipage}{1\linewidth}
			{\includegraphics[width=0.99\linewidth]{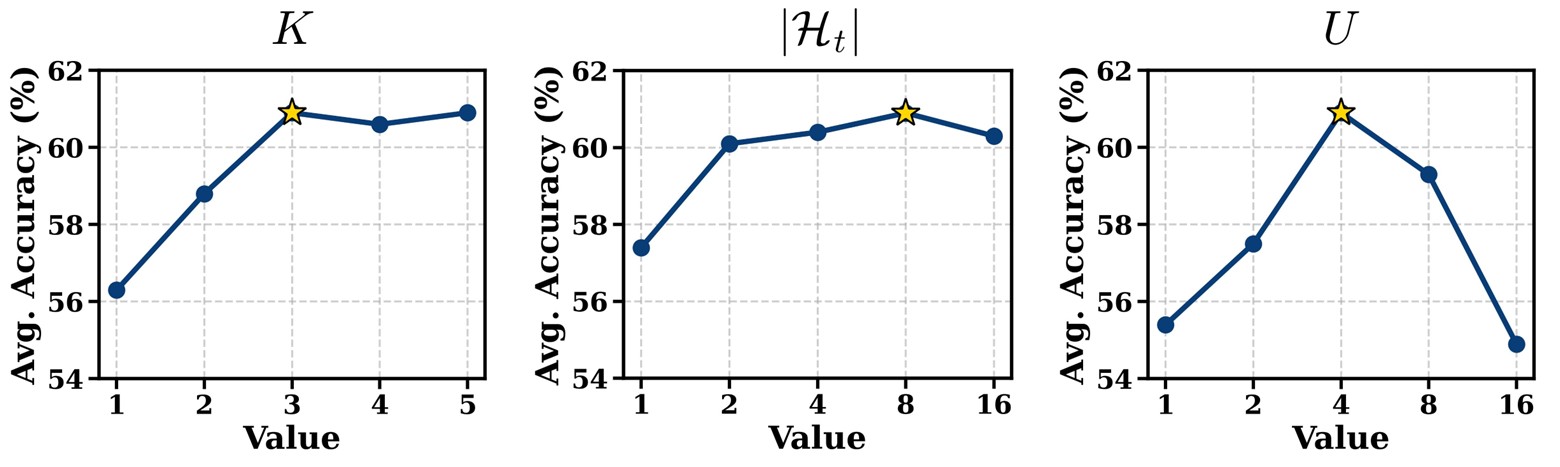}}
			\centering
		\end{minipage}
	\end{center}
\vspace{-6pt}
\caption{Average test performance with GPT-5.4-mini across Spreadsheet, DocVQA, and LiveMath when varying $K$, $|\mathcal{H}_t|$, and $U$. Stars indicate the default settings.}
\label{fig:sensitivity}
\end{figure}

\begin{figure}[t]
\begin{center}
		\begin{minipage}{1\linewidth}
			{\includegraphics[width=0.99\linewidth]{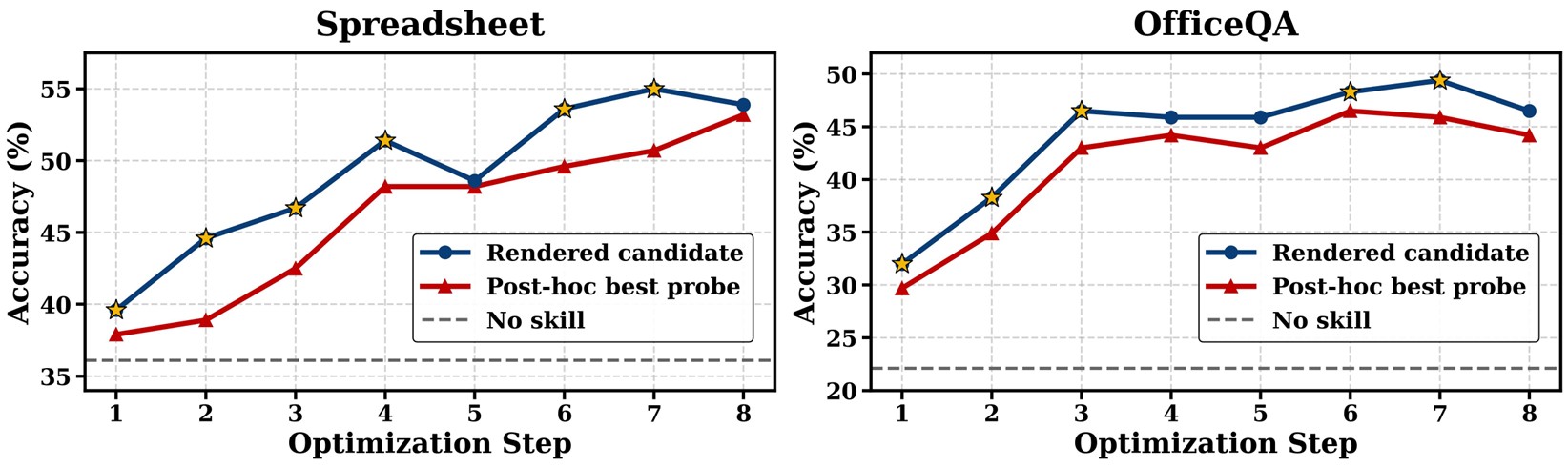}}
			\centering
		\end{minipage}
	\end{center}
\vspace{-6pt}
\caption{Test accuracy of rendered candidates and post-hoc best probes over eight steps on Spreadsheet and OfficeQA with GPT-5.4-mini. Stars mark candidates accepted by the validation gate. The post-hoc best probes are identified using test-set GT only for analysis and never affect optimization.}
\label{fig:dynamics}
\end{figure}  
  
\paragraph{Component Ablations.}
We conduct component ablations on SpreadsheetBench, DocVQA, and LiveMathematicianBench with GPT-5.4-mini as the target model. Direct probe selection replaces the behavior-level pipeline by selecting the probe with the largest training-batch score $\sum_i y_i^k$ and applying the same validation gate. Without behavior clustering, each task-local description forms a separate cluster. Without evidence normalization, SSO uses $e_{ic}^k=y_i^k d_{ic}^k$. Random behavior selection uniformly samples up to $U$ nonzero clusters, while removing the validation gate causes every rendered candidate to be accepted. Table~\ref{tab:ablation} shows that Full SSO performs best. Direct probe selection and random behavior selection cause the largest average drops of 8.5 and 6.1 points, while the other ablations also reduce performance. Additional results are provided in the supplementary material.

\paragraph{Hyperparameter Sensitivity.}
We vary one hyperparameter at a time while fixing the other two. Figure~\ref{fig:sensitivity} shows that the default settings achieve the best performance across all three sweeps, providing a robust configuration for SSO.
  
\subsection{Optimization Dynamics}
Figure~\ref{fig:dynamics} shows that rendered candidates consistently outperform the post-hoc best probes, demonstrating that behavior aggregation and whole-skill rendering can improve beyond individual probes. Accepted updates progressively improve the retained skill, while validation rejects regressions, supporting iterative refinement and the validation gate.

\section{Conclusion}
We formalized GT-free skill optimization and introduced Self-Supervised Skill Optimization (SSO), a comparative framework for learning a reusable skill from unlabeled task instances. SSO compares executions from complete skill probes with corresponding current-skill executions, extracts behavioral differences without seeing the judge's decisions, and aggregates evidence for and against equivalent behaviors. It renders a new complete skill from the highest-ranked behaviors and accepts it only when it wins on a validation set. Across multiple target models, SSO outperforms GT-free prompt optimizers on closed-ended and open-ended tasks, while approaching and sometimes exceeding the strongest GT-based skill optimizer on closed-ended benchmarks.

\bibliography{aaai2027}


\end{document}